\def\year{2017}\relax
\pdfoutput=1

\documentclass[letterpaper]{article} 
\usepackage{aaai17}  
\usepackage{times}  
\usepackage{helvet}  
\usepackage{courier}  
\usepackage{url}  
\usepackage{graphicx}  
\usepackage{times}

\usepackage{graphicx}
\usepackage{amssymb,amsmath,bm}
\usepackage{textcomp}
\usepackage{booktabs}
\usepackage{multirow}
\usepackage[usenames,dvipsnames]{xcolor}
\usepackage{comment}

\usepackage{balance}  
\usepackage{graphics} 
\usepackage{times}    
\usepackage{url}      

\usepackage{xspace}
\usepackage{times}
\usepackage{helvet}
\usepackage{courier}

\usepackage{amsmath}
\usepackage{enumitem}
\usepackage{framed}

 \usepackage{booktabs}
 \usepackage{multirow}

\usepackage{balance}  
\usepackage{graphics} 
\usepackage{times}    
\usepackage{url}      
\usepackage{multicol}

\usepackage{balance}  
\usepackage{booktabs} 
\usepackage{ccicons}  
\usepackage{ragged2e} 
\usepackage{dialogue}
\usepackage[dvipsnames]{xcolor}
\usepackage{CJKutf8}
\usepackage{array}

\newcommand{\lwku}[1]{{\small\textcolor{red}{\bf [#1 --Lwku]}}}

\newcommand{\system}{EmotionPush\xspace}

\newcommand{\SinicaAff}[0]{\ensuremath{1}\xspace}
\newcommand{\CMUAff}[0]{\ensuremath{2}\xspace}
\usepackage{color}
\usepackage{subfig}

\definecolor{angry}{RGB}{247,10,10}
\definecolor{joy}{RGB}{255,255,0}
\definecolor{sadness}{RGB}{40,26,122}
\definecolor{fear}{RGB}{0, 255, 0}
\definecolor{anticipated}{RGB}{255,154,23}
\definecolor{tired}{RGB}{211,43,252}
\definecolor{neutral}{RGB}{128, 128, 128}

\newcommand*\itemw{\item[\textcolor{white}{\begin{CJK*}{UTF8}{gkai}●\end{CJK*}}]}
\newcommand*\itemr{\item[\textcolor{angry}{\begin{CJK*}{UTF8}{gkai}●\end{CJK*}}]}
\newcommand*\itemy{\item[\textcolor{joy}{\begin{CJK*}{UTF8}{gkai}●\end{CJK*}}]}
\newcommand*\itemb{\item[\textcolor{sadness}{\begin{CJK*}{UTF8}{gkai}●\end{CJK*}}]}
\newcommand*\itemg{\item[\textcolor{fear}{\begin{CJK*}{UTF8}{gkai}●\end{CJK*}}]}
\newcommand*\itemo{\item[\textcolor{anticipated}{\begin{CJK*}{UTF8}{gkai}●\end{CJK*}}]}
\newcommand*\itemp{\item[\textcolor{tired}{\begin{CJK*}{UTF8}{gkai}●\end{CJK*}}]}
\newcommand*\itemgrey{\item[\textcolor{neutral}{\begin{CJK*}{UTF8}{gkai}○\end{CJK*}}]}

\newcommand{\cmmnt}[1]{\ignorespaces}

\begin{document}
%

\title{Challenges in Providing Automatic Affective Feedback\\in Instant Messaging Applications}
\vspace{-3pc}
\author{
Chieh-Yang Huang~$^{\SinicaAff}$~~~
Ting-Hao (Kenneth) Huang~$^{\CMUAff}$~~~
Lun-Wei Ku~$^{\SinicaAff}$\\
$^{\SinicaAff}$ Academia Sinica, Taipei, Taiwan. appleternity@iis.sinica.edu.tw, lwku@iis.sinica.edu.tw.\\
$^{\CMUAff}$ Carnegie Mellon University, Pittsburgh, PA, USA. tinghaoh@cs.cmu.edu.
}
\vspace{-3pc}
\maketitle

%
%
%

\begin{abstract}
Instant messaging is one of the major channels of computer mediated communication.
However, humans are known to be very limited in understanding others' emotions via text-based communication.
Aiming on introducing emotion sensing technologies to instant messaging,
we developed \system, a system that automatically detects the emotions of the messages end-users received on Facebook Messenger and provides colored cues on their smartphones accordingly.
We conducted a deployment study with 20 participants during a time span of two weeks.
In this paper, we revealed five challenges, along with examples, that we observed in our study based on both user's feedback and chat logs,
including
{\em (i)} the continuum of emotions, 
{\em (ii)} multi-user conversations, 
{\em (iii)} different dynamics between different users,
{\em (iv)} misclassification of emotions, and
{\em (v)} unconventional content.
We believe this discussion will benefit the future exploration of affective computing for instant messaging, and also shed light on research of conversational emotion sensing.


\end{abstract}

\vspace{-1pc}

\section{Introduction}

\begin{figure}[t]
    \centering
    \hspace{-0.6cm}
    \includegraphics[width=0.5\textwidth]{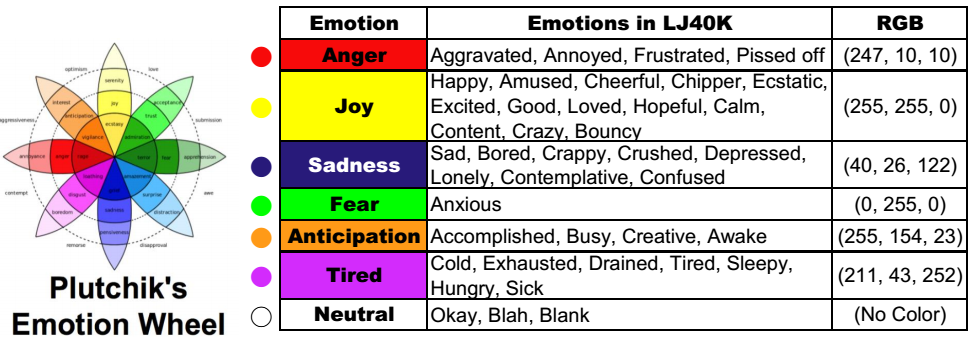}
    \vspace{-1pc}
 	\caption{Visualizing emotion colors with Plutchik's Emotion Wheel. The 40 emotion categories of LJ40K are compacted into 7 main categories, each has a corresponding color on the emotion wheel.}
    \vspace{-1pc}
    \label{fig:color_mapping}
\end{figure}

Text-based emotion detection and classification has a long-lasting history of research~\cite{alm2005emotions}. It is to become increasingly important in the area of machine learning with the increasing emphasis on assistants as frontline interactions for service design. Such assistantship is becoming more manifest in the form of ``chatbots''~\cite{chorusDeploy}, suggesting the research in our work is getting relevant. 
However, compared to content recommendation~\cite{bohus2007olympus,DialPort} or behavioral modeling~\cite{levin2000stochastic}, it is still under discussed. Still less, it has rarely been used in applications for individual users such as instant messengers.
To understand the feasibility of text-based affective computing in the era of mobile devices,
we introduced \emph{\system}\footnote{\system is available at Google Play: \url{https://play.google.com/store/apps/details?id=tw.edu.sinica.iis.emotionpush}}, a mobile application that automatically detects the emotion of the text message that user received via Facebook Messenger,
and provides emotion cues by colors in real-time~\cite{wang2016sensing}.
\system uses 7 colors to represent 7 emotions, which is based on Plutchik's Emotion Wheel color theme (Figure~\ref{fig:color_mapping}.)


For instance, when the user receives a message saying \textit{``Hi, How are you?''},  \system first classifies this message's emotion as \emph{Joy},
and then pushes a notification on the user's smartphone with a yellow icon (Figure~\ref{fig:step_03}), which is the corresponding color of \emph{Joy}.
Later when the user clicks the notification to open Messenger to start the conversation,
\system keeps track on each message that the user receives and uses a color bubble on the top of the screen to continually provide emotion cues (Figure~\ref{fig:step_04}). In Figure~\ref{fig:step_05}, the other party suggests a lunch meeting, which keeps the emotion cue as \emph{Joy}; then the next message about feeling tired changes the emotion cue from \emph{Joy} to \emph{Tired} as in Figure~\ref{fig:step_06}. After giving the last reply which ends this chat session (Figure~\ref{fig:step_07}), users can go back to the desktop but still see the emotion cue of the last message as shown in Figure~\ref{fig:step_08}. Later users can start over and check the notifications from \system again by pulling down the notification bar (Figure~\ref{fig:step_09}).




\newcommand{\widthfactor}[0]{0.21}
\newcommand{\figurefolder}[0]{figure/small_step} 

\begin{figure*}[t]
    \centering
    \subfloat[Receive the first message.\label{fig:step_02}]{%
      \includegraphics[width=\widthfactor\textwidth]{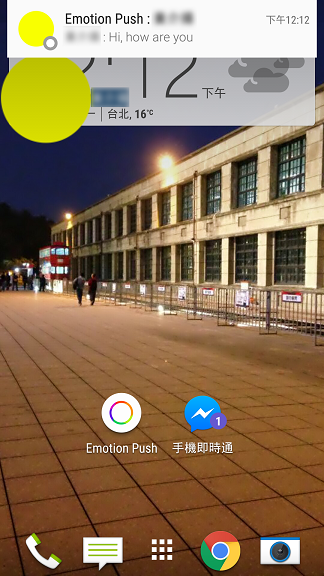}
    }
    \subfloat[Notification.\label{fig:step_03}]{%
      \includegraphics[width=\widthfactor\textwidth]{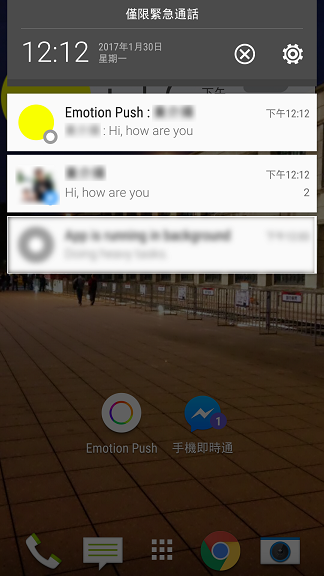}
    }
    \subfloat[Colored bubble.\label{fig:step_04}]{%
      \includegraphics[width=\widthfactor\textwidth]{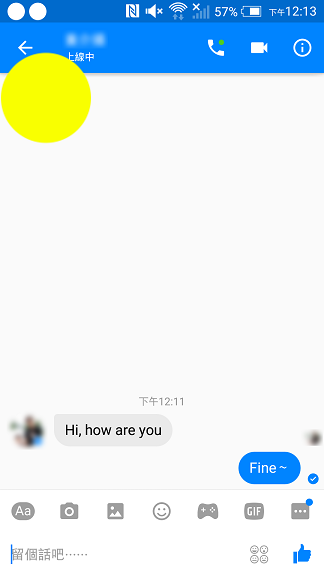}
    }
    \subfloat[Receive a joy message.\label{fig:step_05}]{%
      \includegraphics[width=\widthfactor\textwidth]{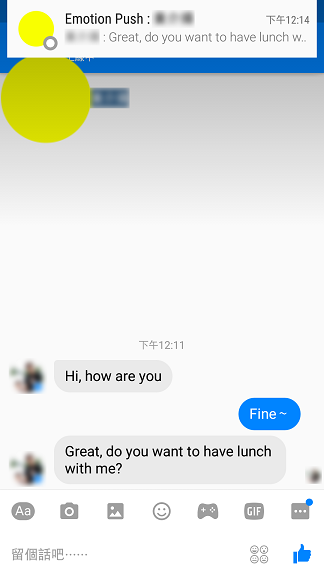}
    }
    \vspace{-1pt}
    \subfloat[Receive a tired message.\label{fig:step_06}]{%
      \includegraphics[width=\widthfactor\textwidth]{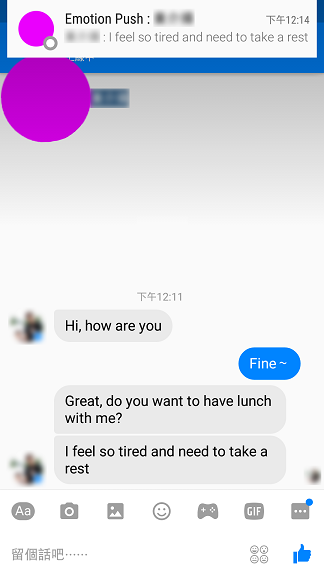}
    }
    \subfloat[Response.\label{fig:step_07}]{%
      \includegraphics[width=\widthfactor\textwidth]{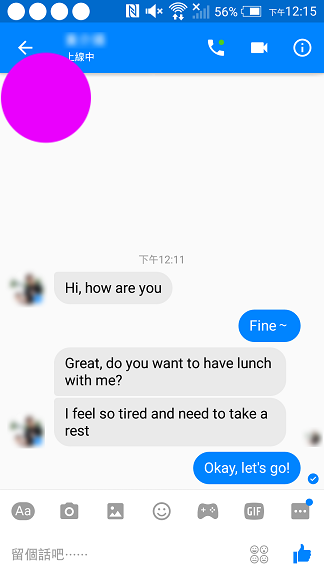}
    }
    \subfloat[Go back to the desktop.\label{fig:step_08}]{%
      \includegraphics[width=\widthfactor\textwidth]{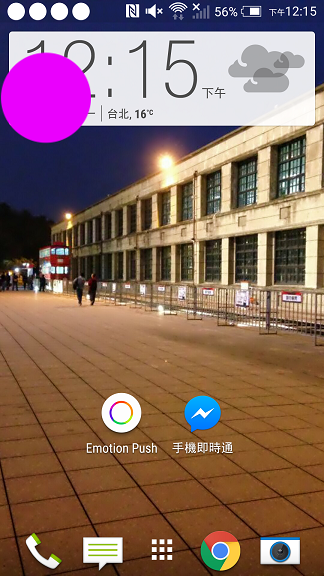}
    }
    \subfloat[Check Notification again.\label{fig:step_09}]{%
      \includegraphics[width=\widthfactor\textwidth]{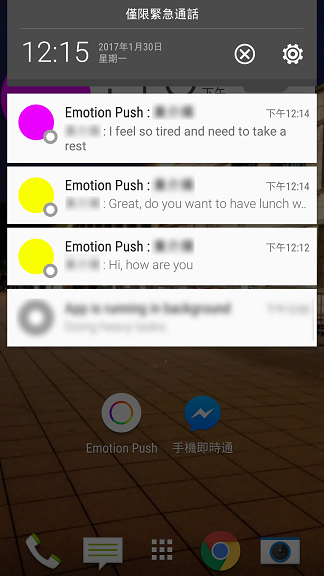}
    }
    \vspace{-1pt}
	\caption{Illustration of using \system.}
	\vspace{-2pt}
    \label{fig:steps}
\end{figure*}

We conducted a deployment study of \system with 20 participants during a time span of two weeks.
20 English native speakers who identified themselves as Messenger frequent users (ages ranged from 18 to 31 years) were recruited.
Participants were asked to install \system on their personal smartphones, keep it open, and actively use it during the whole study.
Each participant was compensated with \$20 US dollars.
In our study, totally 62,378 messages were recorded during 10,623 conversational sessions, which were automatically segmented with 5-minute user timeouts.

In this paper, we first list the potential use cases of EmotionPush.
Then we describe the challenges we identified during the deployment study of \system.
While prior work has shown that identifying emotions based on text is possible, we detail the challenges emerged from the deployment of such a system to the real world.
Challenges that we identified included modeling the continuum of emotional expressions; 
referring the detected emotion to the right speaker in a multi-user conversation;
considering various expression levels per familiarity between users; 
handling the classifier errors;
and problems derived from nonconventional contents such as long messages, code switching, and emojis.

\subsection{EmotionPush System}
\label{subsection:emotion_classification}
The task to predict the emotion of a given text can be regarded as a document (sentence) classification problem if the emotion categories are given. To make sure the developed system achieves good performance, we adopted the powerful classification module LibSVM~\cite{fan2008liblinear}, a supervised-learning tool for Support Vector Machines, and the widely-recognized well-performed feature, Word Embedding, to train the classifier. 

Word Embedding, generated from the neural language models, has been shown to be a successful feature for classification problems~\cite{bengio2006neural,huang2012improving,mikolov2013efficient,mikolov2013distributed}.
We adopted pre-trained 300 dimension word vectors trained on part of Google News\footnote{\url{https://code.google.com/archive/p/word2vec/}} and represented each sentence or document as a vector by averaging all the word vectors of the words in the sentence or the document.

The emotion classifiers of \system were trained on LJ40k~\cite{leshed2006understanding}, a dataset contains 1,000 blog posts for each of 40 common emotions. These classifiers were shown to be comparable to the state-of-the-art methods in terms of performance~\cite{wang2016sensing}. Then \system adopted the categorical representation with the designed color scheme to visualize detected 7 emotions, simplifying the connection between emotions and texts, such that users can easily interact instantly by their instinct. 

\section{Use Case of EmotionPush}

The goal of EmotionPush is to enable end-users to understand better of emotions of their conversational partners.
Hence, the most apparent use case we can think of is to prioritize the messages according to the emotions they convey.
Participants confirmed this use case by indicating that EmotionPush can help them organize their messages, read messages quickly, and managing multiple messages at once.
However, other use cases are not clear before the deployment.
In the use study, participants did name some interesting use cases.
We list these use cases as follows, along with quotes of participants.

\begin{itemize}
  \item \textbf{Emotion Management}
  
  Participants will be able to decide whether they want to receive some information in order to keep their emotion stable. They can either choose to reach the people needed to be taken care, or not to read any message from them to keep themselves neutral:

  \textit{``One of the major advantages was to identify who was angry and needed to talk immediately. That helped me in my interactions a lot.''}
  
  \textit{``If there is a red color chat, I wouldn't read it as it might ruin my mood.''}
  
  \item \textbf{Interacting with People of Little Acquaintance}
  
  Participants mentioned that EmotionPush helps them when talking to strangers or new friends. This makes perfect sense as EmotionPush learns from large datasets and should interpret 
   emotions of messages in a general way. Hence, emotion cues from \system are of valuable reference when we don't know much about the other party.
   
  \item \textbf{Fun Topics to Have}

When participants see some suggested emotions which are different from they expect or interpret, they will confirm with the other party and hence have more topics to create an interesting conversation:
  
  \textit{``It's a funny topic of conversation when the app predicts the emotion interestingly.''}
    
\end{itemize}

Another use case, which is more implicit but draws our attention, is that users may rely on the prompted emotions instead of interpreting received messages by themselves. It is raised by one participant by saying

\textit{``... has the team thought about the social impact this kind of app would have if many people used it? ..., but if everyone were to use an app like this, I feel like people would start to rely on the app too much to determine how they should respond as opposed to figuring out the other person's emotions on their own.''}

However, even with this hidden social concern which should be investigate further, from the result of the user study we can still expect that the system could help people in their interactions from many aspects. The quantitative summary of the user study is reported in Appendix for reference, while the expressed opinions are discussed in this paper. In the following sections, we further detail 5 mentioned challenges emerged from our experiments.

\section{Challenge 1: The Continuum of Emotion}




\system uses a \emph{categorical representation} (e.g. \textit{Anger}, \textit{Joy}, etc.)~\cite{klein2002computer} of emotions instead of a dimensional representation (valence, arousal)~\cite{Sanchez:IHC06} to reduce users' cognitive load.
One natural limitation of applying a categorical representation is the lack of capability of expressing continuum of emotion.
For instance, in the following conversation, the user B sent five consecutive messages, which is less likely to express four different emotions, as predicted\footnote{ \system users do not receive any affective feedback for the messages sent by themselves. In this paper we show colored emotion cues for all messages only for readers' reference. 
The example conversations will be lightly disguised based on the techniques suggested by Bruckman~\cite{bruckman2006teaching} on publication.}:



\begin{itemize}[label={}]
\small
                \itemy \textbf{A:} Aww thanks!!
\vspace{-.4pc}  \itemr \textbf{A:} How's being home?
\vspace{-.4pc}  \itemy \textbf{B:} Studying, haha
\vspace{-.4pc}  \itemb \textbf{B:} But it doesn't feel like I have been away for one year
\vspace{-.4pc}  \itemb \textbf{B:} Nothing has changed here
\vspace{-.4pc}  \itemg \textbf{B:} Time is running so slow now
\vspace{-.4pc}  \itemp \textbf{B:} And I'm still jetlagged, haha
\end{itemize}

While prior work explored modeling continuum of information in text, speech~\cite{yeh2011segment} and video streaming~\cite{gunes2013categorical}, literature had little to say about modeling continuous emotions in a text-based conversation.
To the best of our knowledge, none of the existent conversational datasets contain emotion labels, 
and the continuum property has not been considered in modern emotion labeling systems for conversations.
We believe that considering the hidden emotion states to develop the computational models of humans consecutive dynamics of emotion is a promising direction, where a middle-layered computation which captures the nature flow of emotions is necessary. 






\section{Challenge 2: Multi-User Conversations}




Unexpected challenges were raised by multi-party chatting, which is also known as \textit{Group} or \textit{Channel} in modern messengers.
In our study, in which 22.46\% of messages 
were recorded in multi-user chatting groups, 
we found that providing emotion cues on top of a multi-user conversation would make it difficult for users to concentrate on the running dialog.
For instance, in the following conversation between four different users, it is hard to keep track of both the dialog and the emotion of each user at the same time.

\begin{itemize}[label={}]
\small
                \itemg \textbf{A:} Oh I'll have it tonight, just can't rsvp on mobile arm
\vspace{-.4pc}  \itemgrey \textbf{A:} *atm
\vspace{-.4pc}  \itemgrey \textbf{B:} ACK
\vspace{-.4pc}  \itemb \textbf{B:} I'll mark you down
\vspace{-.4pc}  \itemg \textbf{B:} yup, it's tonight :)
\vspace{-.4pc}  \itemy \textbf{C:} holy shit this sounds awesome!
\vspace{-.4pc}  \itemy \textbf{B:} John \cmmnt{Artemis} is super nerdy rad
\vspace{-.4pc}  \itemg \textbf{D:} I want in on this. I'll see if I can make it work with tech
\vspace{-.4pc}  \itemg \textbf{D:} I can make it
\vspace{-.4pc}  \itemg \textbf{D:} unfortunately my grandparents are coming in tonight so I don't think I'll be able to join :( ha
\end{itemize}

Furthermore, multi-party conversations also raised challenges in designing user experience .
As shown in the Introduction section, \system uses two ways to provide emotion cues: 1) a colored push notification, and 2) a colored bubble that floats on the top layer of the screen.
However, both methods were not capable to efficiently convey emotions in multiple-user conversations.
While a notification can show the message and its emotion cue simultaneously, it only displays the name of the chat group instead of the name of message sender;
users would also find it difficult to identify the corresponding speaker based on bubble's color changes when multiple users are talking.
These design challenges of providing affective feedback that considers emotions, texts and users are beyond prior research of on-line chatting interfaces~\cite{vronay1999alternative,roddy1998interface}.

\section{Challenge 3: Different Dynamics Between Different Users}





Different interaction dynamics occur between people in different context and relationships.
One risk of classifying emotions solely based on texts is the neglect of user context, which is known to have strong correlations with user behavior~\cite{baym2007relational,gilbert2009predicting}.
Prior work has also shown that language comprehension is more than a process of decoding the literal meaning of a speaker’s utterance but making pragmatic inferences that go beyond the linguistic data~\cite{frank2014inferring}.

For instance, in our study, we observed that emotion classification worked better on conversations between users who rarely interacted with each other, in which the languages were more formal.
The following is an example.



\begin{itemize}[label={}]
\small
                \itemgrey \textbf{A:} Hey man.. hows it going!
\vspace{-.4pc}  \itemy \textbf{B:} Hey! It's going well :-)
\vspace{-.4pc}  \itemgrey \textbf{B:} THings are pretty hectic, but I'm trying to get used to the assignments and whatnotn
\vspace{-.4pc}  \itemgrey \textbf{A:} haha sounds like grad school
\vspace{-.4pc}  \itemy \textbf{B:} Yup! Haha
\vspace{-.4pc}  \itemy \textbf{B:} Weren't you planning a trip to Eastern United States \cmmnt{Pittsburgh}?
\vspace{-.4pc}  \itemb \textbf{A:} I was! But I never ended up coming.. I would still like to but my best bet was recruiting and I asked not to go as there was soem work that came up
\end{itemize}

On the other hand, the conversations between users who frequently talked with each other often contain informal expressions. 
The following is an example.

\begin{itemize}[label={}]
\small
\itemgrey \textbf{A:} Okay I was thinking of getting pierced tomorrow after 6:30? I could theoretically do today between like 4:30-6 but I worry about cutting it too close?
\vspace{-.4pc}  \itemr \textbf{B:} I'M DOWN
\vspace{-.4pc}  \itemg \textbf{A:} what time would work best 4 u?
\vspace{-.4pc}  \itemp \textbf{B:} a little after 6:30 might work better bc of activity fair?
\vspace{-.4pc}  \itemy \textbf{A:} Yeah that makes sense!
\vspace{-.4pc}  \itemw [Discussing about inviting other friends]
\vspace{-.4pc}  \itemgrey \textbf{B:} cooooooooooool
\vspace{-.4pc}  \itemgrey \textbf{B:} i can prob get out of helping with teardown haha
\vspace{-.4pc}  \itemg \textbf{A:} Its no big if u cant, its open until 9
\vspace{-.4pc}  \itemgrey \textbf{B:} yeeeee
\end{itemize}

Our observation suggested that \system could be more helpful for some conversations, in this case, the conversations between people who talked less frequently with each other, than for others.
User context could be helpful to both directly improve emotion classification or identify which conversations \system can assist better.

\section{Challenge 4: Misclassification of Emotions}

Emotion classification is not perfect.
It is inevitable that some emotion cues that \system send are incorrect.
For instance, the message of user B in the following conversation should be of \textit{Anticipation} (orange) instead of \textit{Fear} (green).


\begin{itemize}[label={}]
\small
    \itemo \textbf{A:} Will it be factory reset, does it have Microsoft office preset
\vspace{-.4pc}  \itemg \textbf{B:} Yes, I will factory reset it tonight and if you want, you can have a look at it :)
\end{itemize}



In the following example, the message of user A should be of \textit{Anticipation} (orange) instead of \textit{Fear} (green), and the message of user B was apparently not of \textit{Sadness} (blue).


\begin{itemize}[label={}]
\small
                \itemg \textbf{A:} Hey guys so does 2:30 sound good for volunteering tomorrow? We'll take next week off because of fall break
\vspace{-.4pc}  \itemb \textbf{B:} We can leave at 230
\end{itemize}

Misclassified cases raised the questions that what level of performance is good enough for an realistic application.
\system's classifier achieved an average AUC (the area under the receiver operating characteristic curve) of 0.68, which is comparable to the state-of-the-art performance~\cite{wang2016sensing}.
It is noteworthy that humans are not good at identify emotions in text.
Prior work showed that humans on average can only correctly predict 63\% of emotion labels of articles in LiveJournal~\cite{mishne2005experiments}.
Our post-study survey also showed that participants did not think the wrongly-predicted emotion colors are harmful to their chatting experiences (average rating = 0.85, ranges from 0 to 4), while they felt the correctly-predicted emotion colors are helpful (average rating = 2.5).
Given all these factors, we believe that our emotion classifiers' performances are practical for real-world applications.






In addition to improving emotion classification, 
challenges also come from designing good user experience around error cases.
\system is good at identifying \textit{Joy}, \textit{Anger}, and \textit{Sadness}~\cite{wang2016sensing}.
One potential direction is to use different feedback types (e.g., vibration) to distinguish reliable predictions from uncertain ones.






\section{Challenge 5: Unconventional Content}

Similar to most text-processing systems deployed to the real world, \system faced challenges in handling unconventional content in instant messages.
In this section we describe three types of unconventional content we observed in our study: multiple languages, graphic symbols such as emojis, and long messages.

\paragraph{Multiple Languages \& Code Switching}






Real-world users speak various languages.
Even though we recruited English native speakers in our study, participants occurred to speak in, or switch to, various languages when talking with friends.
For example, user A switched between English and Chinese in the following conversation.

\begin{CJK*}{UTF8}{bsmi}
\begin{itemize}[label={}]
\small
                \itemy \textbf{A:} How's ur weekend
\vspace{-.4pc}  \itemp \textbf{A:} Sorry last night I didn't sleep well and needed to work ..Feel like I'm a zombie today haha
\vspace{-.4pc}  \itemgrey \textbf{A:} 整天腦袋空空的
\vspace{-.4pc}  \itemgrey \textbf{A:} 你們都搬到北台灣\cmmnt{台北}？
\vspace{-.4pc}  \itemgrey \textbf{B:} 哈哈加油喔喔喔
\vspace{-.4pc}  \itemgrey \textbf{B:} 對呀!
\vspace{-.4pc}  \itemgrey \textbf{B:} 北海岸\cmmnt{淡水}附近
\vspace{-.4pc}  \itemr \textbf{A:} How r u
\end{itemize}
\end{CJK*}

Not only text-based emotion classification require sufficient labeled data for training, but also code-switching processing techniques relies heavily on training data~\cite{brooke2009cross,vilares2015sentiment}.
All of these technologies are not capable of processing unseen languages.
While prior work explored cross-language acoustic features for emotion recognition in speech~\cite{pell2009recognizing}, detecting emotions in arbitrary languages' texts is still infeasible.
For deployed systems such as \system, making design decisions around languages it can not understand is inevitable.
Currently \system supports two languages, English and Chinese, 
but these two modules were developed separately and still can not handle code-switching case such as the example above.
In the future, we are looking forward to incorporating a language identifier to provide more concrete feedback (e.g., ``Sorry I do not understand French.'') to users.

\paragraph{Emoji, Emoticons, and Stickers}







%
%

\begin{CJK*}{UTF8}{gkai}
Graphic symbols such as emojis, emoticons and stickers are widely used in instant messages, often for expressing emotions.
For example, the emoticon ``\verb|¯\_(ツ)_/¯|'' (also known as ``smugshrug''), 
which represents the face and arms of a smiling person with hands raised in a shrugging gesture, was used in the following conversation.

\begin{itemize}[label={}]
\small
                \itemy \textbf{A:} when can i come and pick up my Jam and also Goat
\vspace{-.4pc}  \itemr \textbf{B:} whenever you want tbh?
\vspace{-.4pc}  \itemr \textbf{B:} we're home rn if ur down
\vspace{-.4pc}  \itemg \textbf{B:} or tomorrow sometime
\vspace{-.4pc}  \itemo \textbf{B:} \verb|¯\_(ツ)_/¯|
\end{itemize}

The usages and effects of graphic symbols in on-line chatting have been thoroughly studied~\cite{jibril2013relevance,walther2001impacts,wang2015more}, 
and techniques of handling emojis in text processing has also been developed~\cite{barbieri2016does,eisner2016emoji2vec}.
However, the current technologies are still not capable to identify emotions from any arbitrary emojis and emoticons,
not to mention new graphic symbols are created everyday (e.g., ``smugshrug'' was just approved as part of Unicode 9.0 in 2016) and stickers are not even text.





\end{CJK*}


\paragraph{Paragraph-like Long Messages}
%






Often instant messaging users chunk a long message into smaller pieces and send them consecutively.
However, we observed that in our study occasionally users send exceptionally long messages. 
For instance, a user sent one message that contains 10 sentences (134 words) to warn the former owner of his/her house to clean up as soon as possible,
a user sent a 10-sentence message (201 words) to advertise his/her incoming stand-up comedy performance, and a user sent a 9-sentence message (152 words) to discuss an reunion event.
In each of these long messages, the user used multiple sentences to express complex issues or emotions, which made it difficult to conclude the message with one single emotion.
While literature showed that emotion classification yielded a better performance on long sentences that contain more words because they bear more information~\cite{calvo2013emotions}, our observation suggested that long messages that contain many sentences often result in a less-confident or incorrect emotion classification as a whole.

\section{Conclusion \& Future Work}
In this paper, we describe challenges in deploying an emotion detection system, \system, for instant messaging applications.
These challenges included the continuum of emotions, multi-user conversations, 
different dynamics between different users,
misclassification, and
unconventional content.
These challenges are not only about providing automatic affective feedback by using text-processing technologies, but also about designing an user experience given the interrelated factors including humanity and languages.
Through these discussions, we expect to gain insight into the deployment of applications of affective computing
and motivate researchers to elaborate the solutions of tomorrow.

In the future,
with the advantage of the developed \system, we plan to design a mechanism which encourages users to contribute their contents and feedback their emotions to advance this technology where it is most needed.


\section{Acknowledgements}
Research of this paper was partially supported by Ministry of Science and Technology, Taiwan, under the contract MOST 104-2221-E-001-024-MY2.

\bibliographystyle{aaai}
\bibliography{emotion_push_final}

\newcolumntype{x}[1]{>{\centering\arraybackslash}p{#1}}
\begin{table*}[ht]
    \centering
    \textbf{{\LARGE Appendix \\~\\}}
    \begin{tabular}{lp{8cm}x{1.3cm}x{1.3cm}x{1.3cm}x{1.3cm}x{1.3cm}}
        \toprule \hline
        &          & \textbf{Disagree} & & & & \textbf{Agree}\\
        & \multicolumn{1}{c}{\textbf{Question}} &  \textbf{1} & \textbf{2} & \textbf{3} & \textbf{4} & \textbf{5} \\ \hline
        1. & In general, I am satisfied with the user experience of \system. & 5\% & 15\% & 20\% & 45\% & 15\% \\
        2. & The mapping between the emotions and the colors are natural to me. & 5\% & 5\% & 50\% & 35\% & 5\% \\
        3. & Changing some of the color might help me to memorize the color mapping. & 5\% & 25\% & 15\% & 45\% & 10\% \\
        4. & \system can predict emotion colors correctly. & 0\% & 40\% & 35\% & 20\% & 5\% \\
        5. & Wrongly predicted emotion colors are harmful to my chatting experiences. & 50\% & 20\% & 25\% & 5\% & 0\% \\
        6. & When predicted correctly, the colors are helpful for me to read and respond messages. & 10\% & 10\% & 30\% & 35\% & 15\% \\
        7. & When facing many messages in notification bar, the correctly predicted colors help me decide which message I should respond firstly. & 20\% & 25\% & 20\% & 30\% & 5\% \\
        8. & When predicted correctly, the colors can enhance social interaction between friends and me in real life. & 10\% & 10\% & 35\% & 30\% & 15\% \\
        9. & When predicted correctly, the colors can decrease my anxiety of misunderstanding/misinterpretation. & 10\% & 10\% & 20\% & 50\% & 10\% \\

        \hline
        10. & Which emotion(s) do you think to be reliable or correct? (*The colors correctly match the emotions) & \multicolumn{5}{p{8cm}}{Joy-Yellow (65\%), Anger-Red (60\%), Sadness-Blue (55\%), Anticipation-Orange (35\%), Tired-Purple (15\%), Fear-Green (15\%), None of them (0\%)} \\
        
        11. & Does \system make you mark messages as read "faster"? Which emotion(s) do those messages belong to? & \multicolumn{5}{p{8cm}}{Anger-Red (40\%), Joy-Yellow (30\%), Sadness-Blue (30\%), None of them (30\%), Anticipation-Orange (15\%), Tired-Purple (15\%), Fear-Green (0\%)} \\
        
        12. & Bearing on. Does \system make you mark messages as read "slower"? Which emotion(s) do those messages belong to? & \multicolumn{5}{p{8cm}}{None of them (40\%), Anger-Red (25\%), Joy-Yellow (20\%), Tired-Purple (20\%), Fear-Green (20\%), Anticipation-Orange (15\%), Sadness-Blue (10\%)} \\
        
        \hline
        13. & Does Emotion Push have other benefits to your social interaction? If so, name some. & \multicolumn{5}{p{8cm}}{
          (As discussed above)
        } \\
        
        14. & In general, what percentage do you use Messenger App / Facebook Webpage to chat with your friends respectively? (ex: 40\% / 60\%) & \multicolumn{5}{p{8cm}}{
            66\% / 34\% (on average)      
        } \\
        
        15. & Which part of Emotion Push do you like the best (the idea, the user interface, and etc.)? & \multicolumn{5}{p{8cm}}{
            idea (80\%) / color coding (15\%) / interface (5\%)
        } \\
        
        16. & How do you think Emotion Push can be improved? & \multicolumn{5}{p{8cm}}{
            (As discussed above)
        } \\
        
        17. & If Facebook Messenger will add emotion color feedback as a new feature in the near future, do you think it's a good idea? & \multicolumn{5}{p{8cm}}{
            Yes (80\%) / No (20\%)
        } \\
        
        18. & Bearing on, Why? & \multicolumn{5}{p{8cm}}{
            (As discussed above)
        } \\
        
        19. & If Facebook Messenger (or any other messaging client such as Line) added the emotion color feedback function, in which scenario you will turn on this function and use it? Please at least name one scenario. & \multicolumn{5}{p{8cm}}{
            turn on (80\%) / turn on if accurate (10\%) / turn off (10\%)
        } \\

        \hline \bottomrule
    \end{tabular}
    \caption{Summary of user study.}
    \label{table:user_study}
\end{table*}

\end{document}